%% file: main.tex
\documentclass[conference]{IEEEtran}

\usepackage{cite}
\usepackage{amsmath,amssymb,amsfonts}
\usepackage{algorithmic}
\usepackage{graphicx}
\usepackage{textcomp}

\usepackage{xcolor}
\usepackage{verbatim}
\usepackage{svg}
\usepackage{eso-pic}
\usepackage[T1]{fontenc}
\usepackage[utf8]{inputenc}

\def\BibTeX{{\rm B\kern-.05em{\sc i\kern-.025em b}\kern-.08em
    T\kern-.1667em\lower.7ex\hbox{E}\kern-.125emX}}
    
\begin{document}

\title{On Safer Reinforcement Learning for \\ Sedation and Analgesia in Intensive Care}

\author{\IEEEauthorblockN{Joel Romero-Hernández}
\IEEEauthorblockA{\textit{BCN MedTech, Complex Systems Lab} \\
\textit{Universitat Pompeu Fabra}\\
Barcelona, Spain  \\
joel.romero@upf.edu}
\and
\IEEEauthorblockN{Oscar Camara}
\IEEEauthorblockA{\textit{BCN MedTech, PhySense Group} \\
\textit{Universitat Pompeu Fabra}\\
Barcelona, Spain \\
oscar.camara@upf.edu}
}

\maketitle

\input{sections/abstract}


\section{Introduction}
\label{section:introduction}
\input{sections/introduction}

\section{Related work}
\label{section:related_work}
\input{sections/related_work}

\section{Background}
\label{section:background}
\input{sections/background}

\section{Methods}
\label{section:methods}
\input{sections/methods}

\section{Results}
\label{section:results}
\input{sections/results}

\section{Conclusion}
\label{section:conclusion}
\input{sections/conclusion}

\section*{Acknowledgment}
\label{section:acknowledgment}
\input{sections/acknowledgment}

\bibliographystyle{IEEEtran}
\bibliography{references}

\end{document}

%% file: sections/abstract.tex
\begin{abstract}
Pain management in intensive care usually involves complex trade-offs, since both inadequate and excessive treatment can compromise patient safety. Prior work on reinforcement learning for sedation and analgesia has explored how to optimize these interventions, but has not considered patient survival or partial observability. To investigate the risks of these design choices, we developed an offline deep reinforcement learning framework that suggests hourly medication doses based on recurrent state representations. Using retrospective data from 47,144 ICU stays in the MIMIC-IV database, we trained and evaluated behavior-regularized actor-critic models that prescribe continuous doses of opioids, propofol, benzodiazepines, and dexmedetomidine according to two goals: reduce pain or jointly reduce pain and 30-day post-discharge mortality. Although the two resulting policies were associated with lower pain, clinician agreement with the pain-only policy was positively correlated with mortality (\(\rho\)=0.119, p<0.0001), while agreement with the joint policy was negatively correlated (\(\rho\)=-0.316, p<0.0001). We found that such divergence arose from a different response to high levels of comorbidity. This suggests that valuing post-discharge outcomes could be critical for learning safer treatment policies, even if a short-term goal remains the primary objective.
\end{abstract}

%% file: sections/introduction.tex
Patients in the intensive care unit (ICU) often experience pain, which can disrupt cardiorespiratory function, lead to emotional distress, and induce long-term sequelae \cite{pain_psychiatric}. Despite the availability of several analgesic and sedative medications, their use can also have profound, even life-threatening effects \cite{pain_opioids_crisis, opioid_overdoses_crisis}. Thus, pain management requires striking a delicate balance between effective pain relief and patient stability \cite{pain_consequences}. 

This is a challenging task, since critically ill patients often have comorbidities that increase vulnerability to side effects \cite{anaesthesia_comorbidities}. Indeed, there are indications that nearly 70\% of patients suffer from unrecognized or undertreated pain \cite{pain_unrecognised}, which increases the likelihood of developing chronic pain: over 100 million Americans suffered from chronic pain in 2010, resulting in an economic loss ranging from \$560 to \$635 billion, higher than that for heart diseases or cancer \cite{pain_chronic_numbers, pain_cost}.

Offline reinforcement learning (RL) can help clinicians address this task by identifying optimal dosing policies from retrospective data \cite{RL_offline_tutorial}. Unlike traditional RL, this avoids the ethical concerns of trial-and-error learning on actual patients.

In sedation and analgesia, data-driven RL has been previously explored, albeit using limited datasets, low-dimensional action spaces, and without accounting for mortality or partial observability \cite{RL_ICU}, both inherent to the ICU. 

We hypothesize that these design choices may yield policies that put patients at risk in exchange for modest short-term gains. Not penalizing mortality, for instance, could create perverse incentives to prioritize actions associated with death after discharge if these are correlated with lower reported pain during the ICU stay. In turn, this behavior could be accentuated by simple action spaces and the absence of strategies to manage imperfect information in sequential decision-making.

\begin{figure}[htbp]
\centerline{
\hspace{0.6cm}
\includegraphics[scale=0.35]{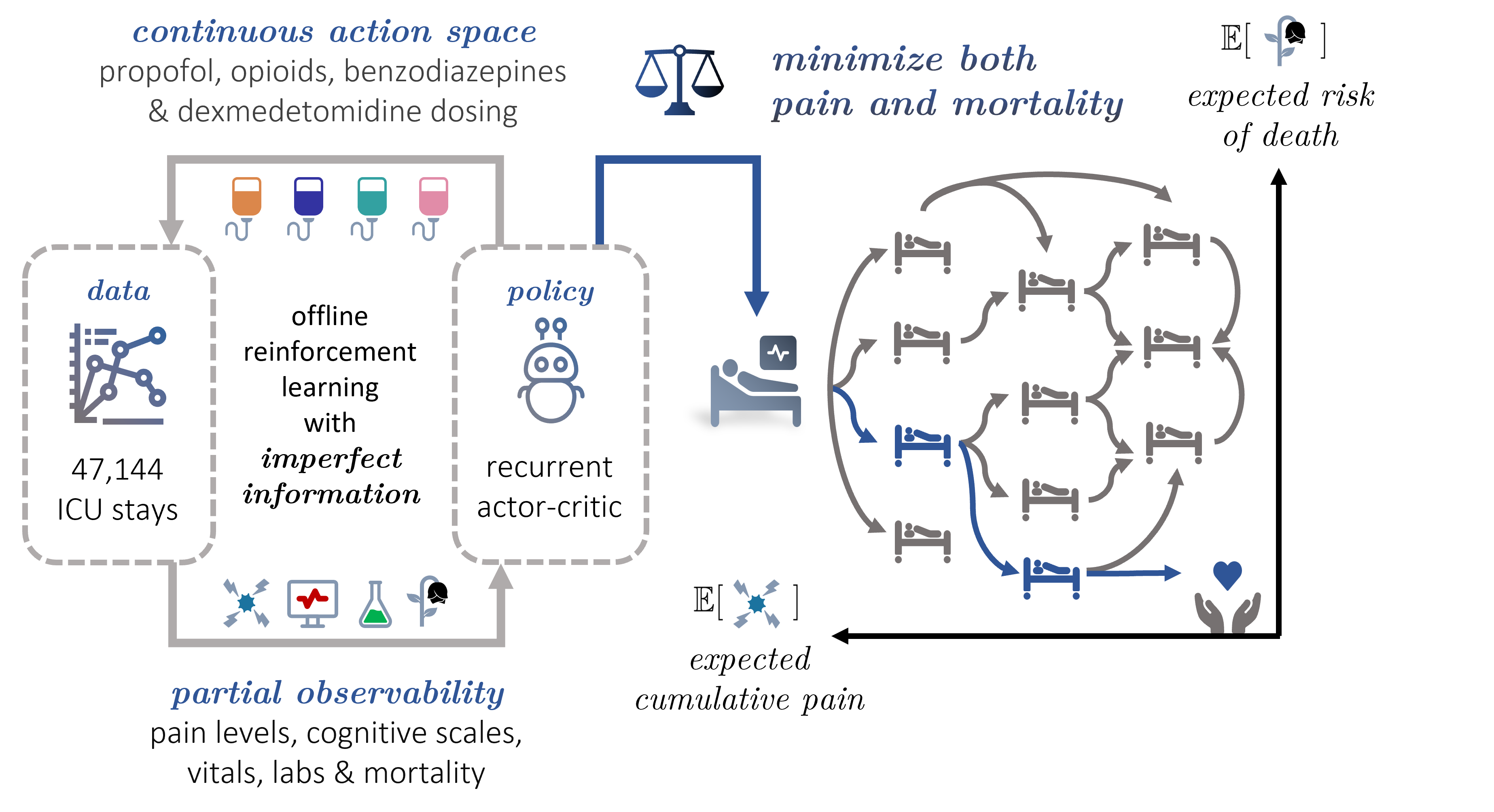}}
\caption{Schematic of our data-driven reinforcement learning (RL) framework for sedation and analgesia in the intensive care unit (ICU). We used offline RL with partial observability to train, validate, and test recurrent agents on 47,144 ICU stays. Specifically, we trained policies to recommend propofol, opioid, benzodiazepine, and dexmedetomidine doses per hour, seeking to minimize pain and mortality. The agents operated based on observations, including vital signs, pain reports, and laboratory values. We studied their behavior to elucidate which objectives encouraged safer strategies.}
\label{fig1}
\end{figure}

As illustrated in Fig. \ref{fig1}, our work addresses these questions through three main contributions. First, we defined reward functions that value both pain relief and survival up to 30 days after discharge. Second, we tackled partial observability and learning without exploration by combining offline RL with learned recurrent representations of the patient state. Third, we developed models based on continuous, real-valued action spaces and a more extensive, highly diverse dataset. 

%% file: sections/related_work.tex
Data-driven RL has been explored in the ICU not only for sedation and analgesia, but also for vasopressor, fluid, insulin, heparin, or ventilator therapy \cite{RL_sepsis, RL_insulin, RL_heparin, RL_ventilator}. These approaches rely on electronic health records (EHRs), which are converted into multivariate time series. Usually, RL agents are trained off-policy to prescribe medications that maximize the reward function, while held-out trajectories are used for offline evaluation. 

Prasad et al. \cite{RL_sedative_ventilator_8_actions_FQI} investigated the optimization of sedation and mechanical ventilation using records of 2,464 successfully discharged patients from MIMIC-III. Fitted Q-iteration (FQI) agents were trained to choose among four sedative doses, using a reward function that valued both vital sign stability and ventilation performance. Similarly, Yu et al. \cite{RL_propofol_ventilator_8_actions_IRL_FQI_GBM, RL_propofol_ventilation_8_actions_SuAC} worked with a subset of 707 cases and used that reward structure to train agents that imitated the clinician policy. 

Lopez-Martinez et al. \cite{RL_morphine_14_actions_D3QN} introduced pain as an observation and proposed dueling double deep Q networks that selected morphine doses from 14 options. This work leveraged a larger dataset by curating records from 6,843 stays in MIMIC-III and employed a reward function that valued pain reduction, heart rate, and respiratory rate stability. 

Eghbali et al. \cite{RL_propofol_fentanyl_DDPG, RL_propofol_midazolam_fentanyl_27_actions_DQN} focused on sedation with two continuous actions: a sedative (propofol) and an opioid analgesic (fentanyl). The authors extracted records from 1,757 adult patients in MIMIC-IV, and a deep deterministic policy gradient agent was trained to optimize mean arterial pressure. They also explored a discrete action space with 27 combinations of dose changes, including propofol, fentanyl, and midazolam.

In summary, previous studies considered at most 6,843 ICU stays, represented as sequences of hourly observations that could include vital signs, ventilation status, sedation, pain levels, medication dosages, and demographics. Only one study explored two continuous actions, while most assumed discretized interventions (with up to 27 options). Some works utilized methods suitable for learning without exploration, such as FQI, yet none explicitly incorporated survival as a constraint or optimization objective, nor modeled the effect of past observations on the patient's current state. 

Building on this prior work, we investigated the effects of penalizing mortality, modeling patient state using recurrent representations, optimizing over four continuous actions, and learning from a larger dataset comprising 47,144 ICU stays.

%% file: sections/background.tex
\textbf{RL.} Sequential decision problems can be modeled as a Markov decision process (MDP) \((\mathcal{S}, \mathcal{A}, p, r, \gamma)\) with state space \(\mathcal{S}\), action space \(\mathcal{A}\), and a state transition function \(p( \mathbf{s}_{t+1}|\mathbf{s}_{t}, \mathbf{a}_{t})\) that follows the Markov property. The reward function \(r\) assigns a scalar \(r_{t}\) to each state-action tuple, and the discount factor \(\gamma \in [0, 1)\) ensures a bounded cumulative reward, the return \(G_{t} = \sum_{k=0}^{\infty} \gamma^k r_{t+k}\). The goal is to learn a policy \(\pi\) that maximizes the expected return \(J(\pi) = \mathbb{E}_{\pi} \left[ G_t \right]\), i.e., an optimal policy \(\pi^{*} = \arg\!\max_{\pi} J(\pi)\). This is often achieved by optimizing action values \(q(\mathbf{s}_{t}, \mathbf{a}_{t}) = \mathbb{E}[G_t | \mathbf{s}_{t}, \mathbf{a}_{t}]\), such that \(q^*(\mathbf{s}_{t}, \mathbf{a}_{t}) = \max_\pi q_\pi(\mathbf{s}_{t}, \mathbf{a}_{t})\) \cite{RL}.

\textbf{Deep RL.} A deterministic continuous control policy can be learned by training a critic network \( Q_{\theta}\) and an actor network \(\pi_{\phi}\), where \(\theta\) and \(\phi\) are their corresponding parameters \cite{DPG}. Here, \( Q_{\theta}\) learns to estimate \(q_{\pi_{\phi}}(\mathbf{s}_t, \mathbf{a}_t)\) by minimizing the error between the value predicted for \(\pi_{\phi}\)'s actions and Bellman targets \(y_t = r_t + \gamma Q_{\theta'}(\mathbf{s}_{t+1}, \pi_{\phi'}(\mathbf{s}_{t+1}))\). Concomitantly, \(\pi_{\phi}\) is trained to maximize \(J(\phi) = \mathbb{E}[Q_{\theta}(\mathbf{s}_t, \pi_{\phi}(\mathbf{s}_t))]\). Here, \(Q_{\theta'}\) and \(\pi_{\phi'}\) are target networks, slowly updated through Polyak averaging to stabilize the learning process \cite{DDPG}.

\textbf{Offline learning.} Training \(Q_{\theta}\) from experiences collected by an arbitrary policy involves an important extrapolation error: incomplete coverage of the state-action space leads to biased value estimates and underperforming policies \cite{BCQ}, which are compounded by the use of function approximation \cite{overestimation}. Offline RL introduces strategies to minimize the distributional shift between the target and behavior policy, decreasing the influence of these out-of-distribution actions \cite{BEAR, CQL}.

\textbf{Partial observability.} A partially observable Markov decision process (POMDP) \((\mathcal{S}, \mathcal{A}, \mathcal{O}, p, e, r, \gamma )\) models imperfect information as a space of observations \(\mathcal{O}\) \cite{stochastic_POMDP}. In a POMDP, the emission function \(e\) maps each hidden state \(\mathbf{s}_t\) to an observation \(\mathbf{o}_t \in \mathcal{O}\). Because \(\mathbf{s}_{t+1}\) depends on \(\mathbf{s}_t\), observations do not satisfy the Markov property. Instead, a sufficient statistic can be constructed from the current history of observations and actions, \(\mathbf{h}_{t} = (\mathbf{o}_{0}, \mathbf{a}_{0}, \mathbf{o}_{1}, \mathbf{a}_{1},  ..., \mathbf{o}_{t-1}, \mathbf{a}_{t-1}, \mathbf{o}_{t})\).

%% file: sections/methods.tex
\subsection{Formalization}

\textbf{Model.} We formulated pain management as a POMDP where \(\mathbf{s}_t \in \mathcal{S} \subseteq \mathbb{R}^{d_s}\) is the hidden, \(d_s\)-dimensional state of the patient, \( \mathbf{a}_t \in \mathcal{A} \subseteq \mathbb{R}^{d_a}\) is a vector of doses administered for \(d_{a}\) medications of interest, and \(\mathbf{o}_t \in \mathcal{O} \subseteq \mathbb{R}^{d_o}\) is a vector of \(d_{o}\) observations taken from the patient. The state transition function \(p\) captures the unknown mechanics of the patient system, and the emission function \(e\) represents the processes mapping these latent dynamics to measurements. Finally, the reward function \(r\) and the discount factor \(\gamma\) encode the objectives and priorities for some dosing policy \(\pi\).

\textbf{Trajectories.} A trajectory \(\boldsymbol{\tau}^{i}\) describes the evolution of an ICU stay \(i\) according to the POMDP. It includes a sequence of \(T_{i}\) observation-action tuples \(\smash{(\mathbf{o}_{t}^{i}, \mathbf{a}_{t}^{i})}\), and a binary outcome \(m^{i}\), with \(m^{i}=0\) if the patient survived and \(m^{i}=1\) if the patient died. Every measurement \(\mathbf{o}_{t}^{i}\) contains a dimension \(\varsigma^{i}_{t} \in \{0, 1, ..., \varsigma_{\text{max}}\}\) encoding the patient's reported pain for time step \(t\). Hence, each trajectory is defined as a tuple \(\boldsymbol{\tau}^{i} = \big((\mathbf{o}_{t}^{i}, \mathbf{a}_{t}^{i})_{t=0}^{T_i-1}, m^{i}\big)\) with a reward sequence \(\big(r_{t}^{i})_{t=0}^{T_i-1}\), where \(r_{T_i-1}^{i}\) can include an additional terminal reward.

\textbf{Policy.} We define the reward function \(r\) to score the utility of sedative and analgesic dosing decisions. We aim to learn policies \(\pi\) that maximize \(\mathbb{E}[G_t]\) from a dataset \(\mathcal{D}\) containing \(N\) pre-recorded trajectories. To approximate a sufficient statistic for decision-making under partial observability, we define pain management policies as maps \(\pi: \mathcal{H} \to \mathcal{A}\) from history vectors \(\mathbf{h}^i_t\in \mathcal{H}\) to action vectors \(\mathbf{a}^i_t\). Here, \(\mathcal{H}\) is the space of possible histories and \(\mathbf{a}^i_t\) encodes the \(d_{a}\) doses administered to the patient at time step \(t\) for trajectory \(i\).

\textbf{Reward.} Equation \eqref{tab:eq:reward} encodes our general reward structure, which includes two weighted terms. The first term penalizes higher levels of pain per time step, normalized by stay length and scaled by weight \(w_{\varsigma}\). The second term, scaled by weight \(w_m\), introduces an explicit terminal penalty for mortality, which is added to the reward at the last time step:
\begin{equation}
r_t^i =-w_{\varsigma} \cdot \frac{1}{T_i} \cdot \frac{\varsigma_t^i}{\varsigma_{\max}}-
w_m \cdot \mathbb{I}_{\{t = T_i-1\}} \cdot m^i.
\label{tab:eq:reward}
\end{equation}
Thus, the relative importance of the two sub-goals for a given policy can be determined by simply adjusting the \(w_{m}/w_{\varsigma}\) ratio.

\subsection{Data engineering}
\begin{figure}[htbp]
\centerline{\hspace{-0.25cm} \includegraphics[scale=0.28]{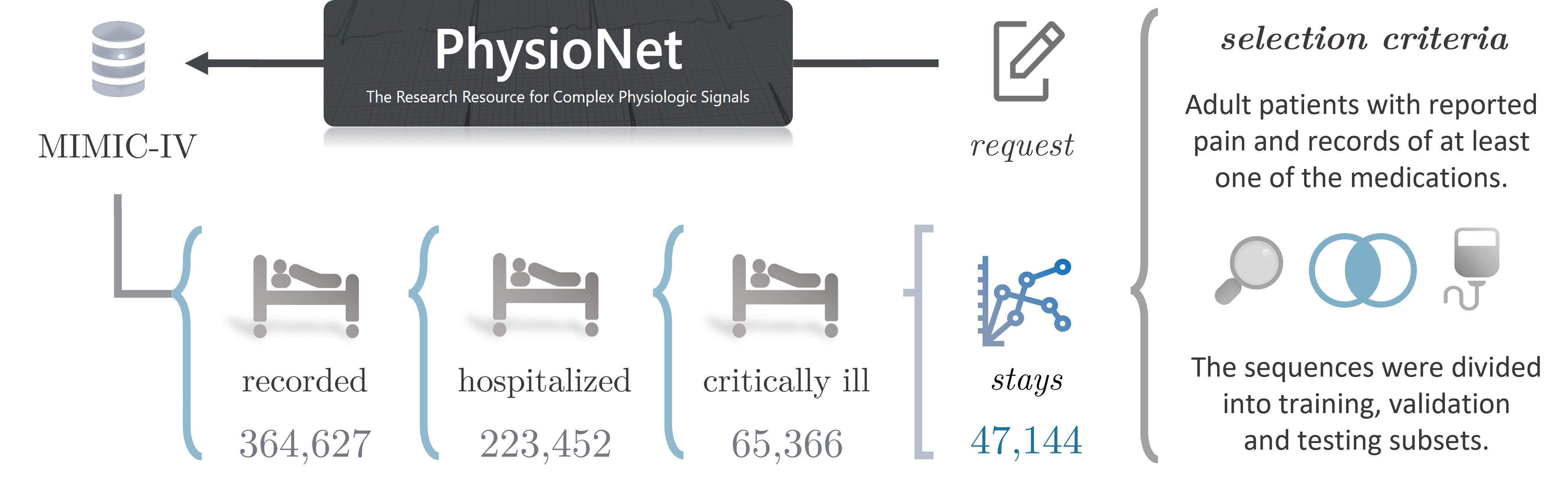}}
\vspace{0.2cm}
\centerline{\includegraphics[scale=0.26]{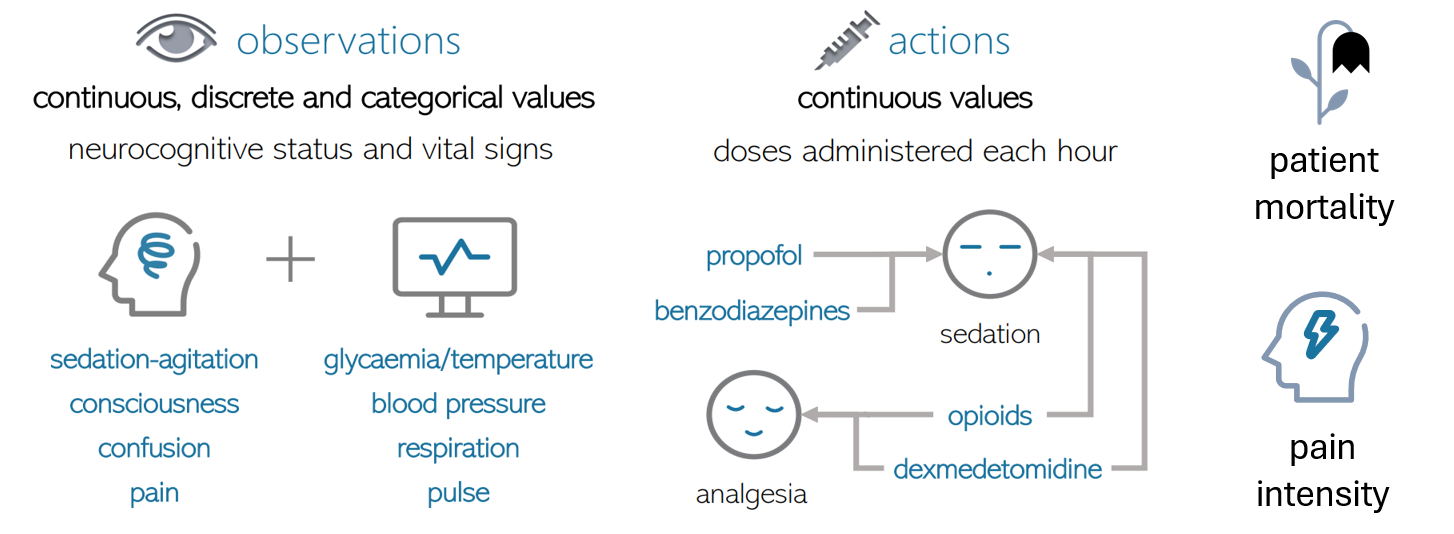}}
\caption{Dataset preparation including quantitative and ordinal observations, apart from four continuous medication actions per hour.}
\label{fig2}
\end{figure}

As shown in Fig. \ref{fig2}, the MIMIC-IV database \cite{mimic4} was used to curate a dataset with the records of 47,144 stays from 30,087 adult patients. Each stay trajectory was encoded as a 22-dimensional hourly time series, comprising both quantitative (e.g., blood pressure) and ordinal (e.g., reported pain) features, as well as a binary survival marker (30 days post-discharge).

\textbf{Observations.} The observation space \(\mathcal{O}\) included vital signs (heart rate, respiratory rate, pulse oxygen saturation, mean, systolic, and diastolic blood pressure), body temperature, urine output, laboratory values (glucose, potassium, sodium, chloride, bicarbonate, creatinine, and blood urea nitrogen), and cognitive assessment scales (pain intensity reports, Glasgow Coma Scale scores, sedation-agitation, and delirium).

\textbf{Actions.} The action space \(\mathcal{A}\) incorporated four continuous dimensions populated with dosing records from eight sedatives and analgesics. Firstly, we expressed opioid administrations as a combined dose per hour by converting fentanyl and hydromorphone doses into oral morphine-equivalent mg \cite{opioid_conversion} and merging these with pre-existing morphine records. Secondly, we normalized propofol doses into mg per kg  \cite{propofol_doses}. Thirdly, we reported benzodiazepine doses as midazolam-equivalent mg by converting diazepam records into lorazepam-equivalent mg \cite{diazepam_lorazepam}, merging these with lorazepam doses, and then converting to midazolam-equivalent mg \cite{lorazepam_midazolam}. Lastly, dexmedetomidine doses were normalized to mg per kg \cite{dexmedetomidine_doses}. 

\textbf{Outlier removal.} As proposed in Gupta et al. \cite{preprocessing}, quantile-based data cleaning was applied to continuous variables. Thresholds were tuned in the training data to exclude physiologically implausible values. For vital signs, values below the 0.05th percentile and above the 99.95th were deleted. For laboratory values, the thresholds were 0.1 and 99.9. For urine output and medication doses, which cannot be negative, values beyond the 99.9th and the 99th percentile were deleted.

\textbf{Imputation.} To fill missing values, we first carried the last observation forward according to feature-specific clinical thresholds \cite{sample_and_hold}. Then, we implemented a pipeline based on multiple imputation by chained equations (MICE) to iteratively estimate the remaining missing values \cite{mice}. We chose gradient boosting machines (GBMs) with decision trees (DTs) \cite{gbm} as our predictive model for two reasons. On the one hand, DTs are well-suited for capturing intricate non-linear relationships among both quantitative and categorical variables \cite{decision_trees}. On the other hand, DT ensembles have been reported to achieve state-of-the-art performance in complex tasks \cite{xgboost, xgboost_clinical}. 

\subsection{Architecture and algorithm}

\begin{figure}[htbp]
\vspace{-1mm}
\centerline{\includegraphics[scale=0.3]{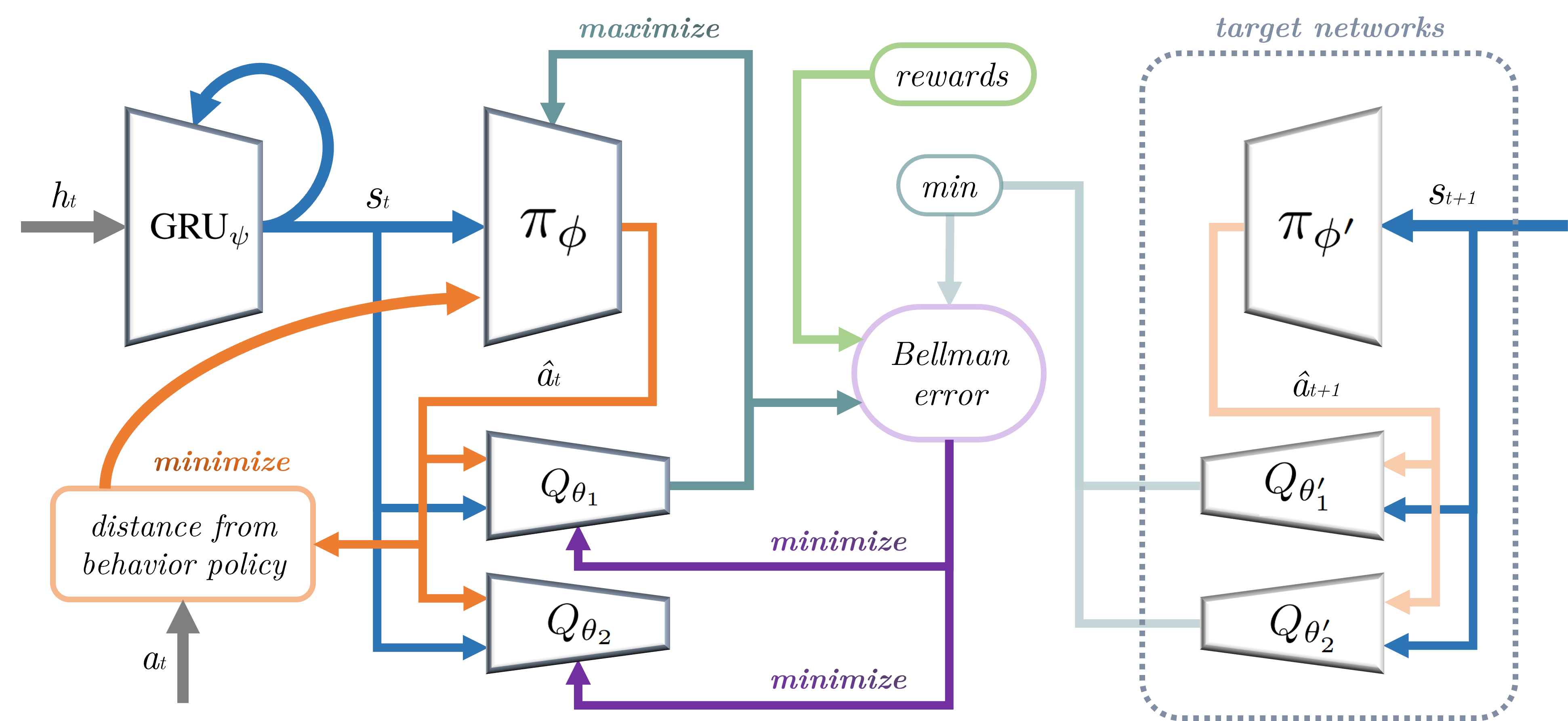}}
\caption{Actor-critic architecture to address both imperfect information and offline learning. The gated recurrent unit network \(\text{GRU}_{\psi}\) learns to capture the influence of past observations in history \(\mathbf{h}_t\) to approximate the patient state \(\mathbf{s}_t\). Then, the behavior-regularized actor \(\pi_{\phi}\)  proposes actions \(\mathbf{a}_t\) based on those states, which are evaluated by the critics \(Q_{\theta_1}\) and \(Q_{\theta_2}\).}
\label{fig3}
\end{figure}

To learn our policies, we implemented the three-module architecture shown in Fig. \ref{fig3}. First, a recurrent neural network with gated recurrent units (GRU) \cite{GRU} and parameters \(\psi\), \(\text{GRU}_{\psi}\). Second, two critic networks \(Q_{\theta_1}\) and \(Q_{\theta_2}\), with target networks \(Q_{\theta_1'}\) and \(Q_{\theta_2'}\). Third, an actor network \(\pi_{\phi}\) with target network \(\pi_{\phi'}\). As discussed below, this architecture was intended to mitigate both extrapolation error and state aliasing.

\textbf{State representation.} \(\text{GRU}_{\psi}\) was trained to approximate an \(\mathcal{H} \to \mathcal{S}\) map by learning compact representations from histories of observations and actions \cite{recurrent_DPG_SVG}.  As shown in \eqref{eq:gru_loss}, this was done in a supervised manner, focusing on predicting both future observation vectors \(\mathbf{o}^i_{t+1}\) and mortality \(m^i\), where \(\hat{m}^i_t\) is the death probability predicted at time \(t\):
\begin{equation}
\mathcal{L}_{\text{GRU}}(\psi) = \left\lVert \mathbf{o}^{i}_{t+1} - \hat{\mathbf{o}}^{i}_{t+1} \right\rVert_2^2 + 0.2 \cdot \ell_m(m^i, \hat{m}^i_t),
\label{eq:gru_loss}
\end{equation}
\begin{equation}
\ell_m(m^i, \hat{m}^i_t) = - m^i \log(\hat{m}^i_t) - (1 - m^i)\log(1 - \hat{m}^i_t).
\label{eq:m_loss}
\end{equation}
The mortality term \eqref{eq:m_loss} was down-scaled by 0.2 to encourage representations that captured both relevant dynamics and long-term outcomes according to validation performance.

\textbf{Double critic.} We trained \(Q_{\theta_1}\) and \(Q_{\theta_2}\) to predict the expected cumulative reward for medication doses based on our state representations. That is, to approximate maps \(\mathcal{S} \times \mathcal{A} \to \mathbb{R}\) by estimating the return \(G_t\) given a reward function \(r\). Each critic consisted of a multi-layer perceptron (MLP) trained to minimize the mean squared error between its action-value estimates and Bellman targets \(y^i_t\):
\begin{equation}
\mathcal{L}_{Q}(\theta_j) = \mathbb{E}_{\mathcal{D}}
\left[\left(Q_{\theta_j}(\mathbf{s}^i_t, \mathbf{a}^i_t) - y^i_t
\right)^2 \right] \quad j \in \{1,2\}.
\end{equation}
The targets were computed by selecting the lowest prediction between the target networks \(Q_{\theta'_1}\) and \(Q_{\theta'_2}\),
\begin{equation}
    y^i_t = r^i_t + \gamma \min_{j=1, 2} Q_{\theta'_j}(\mathbf{s}^i_{t+1}, \pi_{\phi'}(\mathbf{s}^i_{t+1})).
\label{eq:targets}
\end{equation}
By taking the minimum of the two estimates, a conservative bias is introduced. This pessimistic value estimation process has been proven to mitigate the overestimation bias in value-based RL methods \cite{TD3} while maintaining a relatively low computational overhead compared to larger critic ensembles.

\textbf{Behavior-regularized actor.} We trained \(\pi_{\phi}\) (also an MLP) to learn a mapping \(\mathcal{S} \to \mathcal{A}\) that maximized the estimates of the first critic, \(Q_{\theta_1}\). Following prior work on minimalist offline RL \cite{TD3_BC}, actor updates were regularized with a behavior cloning objective that penalized deviation from clinician actions:
\begin{equation}
\mathcal{L}_{\pi}(\phi) = \mathbb{E}_{\mathcal{D}} \Big[-\lambda \, Q_{\theta_1}\big(\mathbf{s}^i_{t},\pi_{\phi}(\mathbf{s}^i_{t})\big)
+\|\pi_{\phi}(\mathbf{s}^i_{t})-\mathbf{a}^i_{t}\|_2^2\Big].
\label{eq:regularized_actor_loss}
\end{equation}
The first term in \eqref{eq:regularized_actor_loss} encourages the policy to select actions with high estimated value, while the second term mitigates distributional shift by constraining the policy to remain close to actions observed in the dataset. Here, \(\lambda\) acts as an adaptive normalization for critic values that ensures scale consistency between both terms. In this context, the importance of value maximization relative to behavior cloning can be tuned by increasing the hyperparameter \(\alpha\), such that
\begin{equation}
\lambda = \frac{\alpha}{\mathbb{E}_{\mathcal{D}}\big[|Q_{\theta_1}(\mathbf{s}^i_{t},\mathbf{a}^i_{t})|\big]}.
\end{equation}

\subsection{Experiments}
\noindent We learned two distinct policies. In both cases, we set \(w_{\varsigma} > 0\).

\textbf{Policy A}. We set \(w_{m} = 0\) to train a policy \(\pi_{\phi_{A}}\) that only optimized for pain minimization. This was intended to model a strategy prioritizing short-term effectiveness:

\small
\begin{equation}
\phi_A \in \arg\!\min_{\phi} \Bigg\{
\mathbb{E}_{\tau \sim \pi_{\phi}}
\left[ 
    \frac{w_{\varsigma}}{T}\sum_{t=0}^{T-1}
    \frac{\varsigma_t}{\varsigma_{\max}}
\right]
\Bigg\}.
\end{equation}
\normalsize

\textbf{Policy B}. We enforced \(w_m > w_{\varsigma}\) to train a policy \(\pi_{\phi_{B}}\) that jointly minimized reported pain and mortality up to 30 days after discharge. Since both terms in \eqref{eq:objective_b} are normalized to lie in \([0,1]\), this ensures that a full-unit increase in mortality incurs a larger penalty than any possible improvement in the pain term, while keeping both on a comparable scale. This models a safety-aware variation of the treatment strategy, which prioritizes patient survival:

\small
\begin{equation}
\phi_B \in \arg\!\min_{\phi} \Bigg\{ \; \mathbb{E}_{\tau \sim \pi_{\phi}}
\left[
\, \frac{ w_{\varsigma}}{T}\sum_{t=0}^{T-1} \frac{\varsigma_t}{\varsigma_{\max}} +\; 
w_m\,m
\right]  
\Bigg\}.
\label{eq:objective_b}
\end{equation}
\normalsize

\textbf{Data splits.} 64\% of the ICU stays were used to train our representations and offline RL policies, while 16\% were left for internal validation and model selection. Conversely, 20\% of the stays were excluded from the learning process and reserved for testing and policy analysis. This split was performed while ensuring that no patient had stays in multiple subsets. 

\textbf{Hyperparameters.} We studied the effect of different latent dimensions, types of recurrent units, activation functions, weight initialization methods, learning rates, optimizers, and goal preference weights. The final settings shown in Table \ref{tab:network_hyperparameters} were chosen based on the state encoder validation loss and the Bellman error of the critic networks on the validation subset. For policy B, the selected preference ratio was \(w_m = 10 \cdot w_{\varsigma}\).

\begin{table}[htbp]
\caption{Neural network architecture and hyperparameters.}
\begin{center}
\setlength{\tabcolsep}{2pt}
\begin{tabular}{lccc}
\hline
\textbf{Component} & \textbf{State Encoder} & \textbf{Critic} & \textbf{Actor} \\
\hline
Input dimensionality 
& \(d_o + d_a\)
& \(d_s + d_a\)
& \(d_s\) \\
Output dimensionality 
& \(d_s\) 
& \(1\) 
& \(d_a\) \\
Latent dimensions
& \(64\) 
& \(64\) 
& \(64\) \\
Hidden layers 
& \(2\) 
& \(2\) 
& \(2\) \\
Activation function 
& \(\tanh\) 
& LeakyReLU 
& LeakyReLU \\
Weight initialization 
& Orthogonal 
& Orthogonal 
& Orthogonal \\
Base architecture
& GRU 
& MLP 
& MLP \\
\(\alpha\)
& -- 
& --
& \(2.0\) \\
\(\gamma\)
& --
& \(0.99\)
& -- \\
Polyak coefficient
& -- 
& \(0.005\) 
& \(0.005\) \\
Optimizer 
& Adam 
& Adam 
& Adam \\
Learning rate 
& \(1 \times 10^{-4}\) 
& \(1 \times 10^{-4}\) 
& \(1 \times 10^{-4}\) \\
Gradient clipping 
& \(0.5\) (norm) 
& \(0.5\) (norm) 
& \(0.5\) (norm) \\
\hline
\end{tabular}
\label{tab:network_hyperparameters}
\vspace{-0.25cm}
\end{center}
\end{table}

\subsection{Policy analysis}
In the absence of a provably accurate counterfactual model, we grounded our evaluation in simple, interpretable metrics based on the empirical evidence in the data distribution. For each learned policy, we computed a clinician-agent agreement score \(C_{i}\) for every trajectory \(i\) in the test set:
\begin{equation}
C_{i}=1-\frac{1}{d_a \cdot T_i}\sum_{k=1}^{d_a}\sum_{t=0}^{T_i-1}\frac{\left|a^{i,k}_{t}-\pi_{\phi}(\text{GRU}_{\psi}(\mathbf{h}^i_{t}))^{k}\right|}{a^{k}_{\max}}.
\label{eq:agreement}
\end{equation}
As defined in \eqref{eq:agreement}, the agreement score corresponds to one minus the normalized mean absolute distance between the recorded clinician doses \(\mathbf{a}^{i}_{t}\) and the proposed doses \(\pi_{\phi}(\text{GRU}_{\psi}(\mathbf{h}^i_{t}))\). Here, \(d_a\) is the number of medications, \(a^{i,k}_{t}\) is the dose recorded for action \(k\) at time \(t\), and \(\smash{a^{k}_{\max}}\) refers to the maximum dose recorded for that medication. 

We used this score to measure the overall similarity between the actions proposed by our policies and the clinicians' decisions. Then, we identified the cases where the deviation was greatest. To better understand these differences, we extracted post-discharge diagnostic data and identified the comorbidity profiles associated with the most salient behaviors. 

Subsequently, we examined associations between clinician-agent agreement, mortality up to 30 days after discharge, and cumulative pain by estimating bootstrapped Spearman rank correlations (1,000 resamples) in test cases. This correlation metric was chosen to avoid assumptions of linearity or scale, providing a more robust basis for policy comparison.

This aimed to assess how closely each policy aligned with clinical strategies associated with specific outcomes. In addition, we evaluated the contribution of individual medications to the overall correlation. We also estimated bootstrapped correlations between patient comorbidity scores and deviation from doses prescribed by clinicians for specific medications.

%% file: sections/results.tex
\begin{figure}[htbp]
\centerline{
\vspace{-0.1cm}
\hspace{-0.5cm}
\includesvg[scale=0.52]{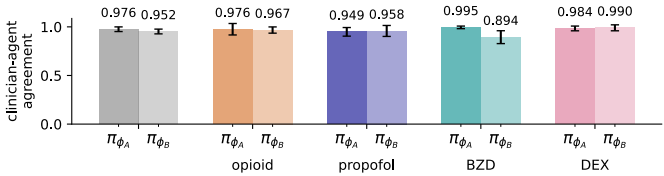}
}
\caption{Clinician-agent agreement for policies A (\(\pi_{\phi_A}\)) and B (\(\pi_{\phi_B}\)) in the test set. Bars show the mean similarity, overall and medication-specific, including opioids, propofol, benzodiazepines (BZD), and dexmedetomidine (DEX). Error bars indicate the standard deviation across patient stays.}
\label{fig4}
\end{figure}

Policy A (\(\pi_{\phi_A}\)) and policy B (\(\pi_{\phi_B}\))  exhibited high aggregate agreement with the doses recorded in test cases (Fig. \ref{fig4}). This implied that the behavior cloning objective constrained the training process effectively: on average, both actors learned to operate within the ranges expected for typical clinician actions. 

\begin{figure}[htbp]
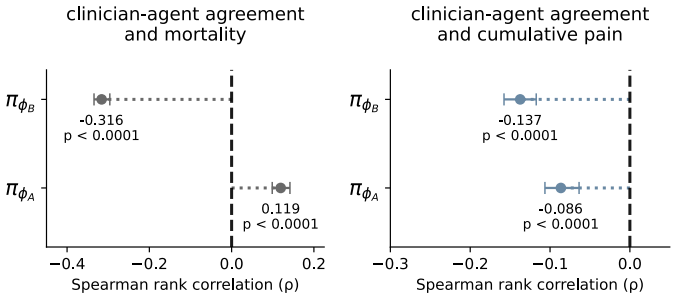

\hspace{-0.2cm}
\centerline{
\includesvg[scale=0.57]{figures/policy_and_mortality}
\hspace{-0.38cm}
\includesvg[scale=0.57]{figures/policy_and_pain}
}
\caption{Correlation between clinician-agent agreement and mortality (left) or cumulative pain (right) for policy A (\(\pi_{\phi_A}\)) and policy B (\(\pi_{\phi_B}\)). Points are correlation estimates, and horizontal bars represent confidence intervals. The vertical dashed lines denote no association (\(\rho\) = 0).
}
\label{fig5}
\end{figure}

Nevertheless, when examining the association between clinician-agent agreement, mortality, and cumulative pain, we observed marked differences between the two policies. As shown in Fig. \ref{fig5}, higher clinician-agent agreement with \(\pi_{\phi_A}\) was significantly associated with lower reported cumulative pain, but also with higher mortality up to 30 days after discharge. In contrast, agreement with \(\pi_{\phi_B}\) was significantly and negatively correlated with mortality, while also exhibiting a more pronounced negative correlation with cumulative pain. 

Fig.~\ref{fig6} examines this pattern in greater detail for \(\pi_{\phi_B}\). Despite general broad support for the action distribution proposed by this policy, increasing clinician-agent agreement was associated with progressively lower empirical mortality rates and cumulative levels of reported pain.

\begin{figure}[htbp]
\centerline{
\includesvg[scale=0.52]{figures/B_and_mortality}
\includesvg[scale=0.52]{figures/B_and_pain}
}
\caption{Association between clinician-agent agreement, mortality, and cumulative pain for policy B. Points represent agreement bins, with color intensity indicating the number of cases per bin. For mortality (left), the vertical axis reports the empirical mortality rates, that is, the fraction of patients who died for each agreement score. For cumulative pain (right), points denote the mean cumulative pain per bin, while error bars indicate the standard deviation.
}
\label{fig6}
\vspace{-5mm}
\end{figure}
\begin{figure}[htbp]
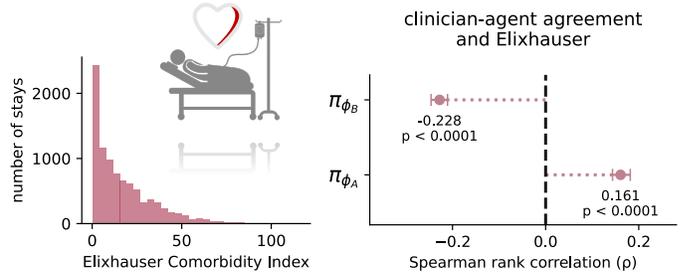

\vspace{-0.2cm}
\hspace{-0.2cm}

\centerline{
\includesvg[scale=0.12]{figures/elixhauser}
\hspace{-0.45cm}
\includesvg[scale=0.57]{figures/policy_and_elixhauser}
}
\caption{Elixhauser Comorbidity Index (ECI) distribution (left) and correlations between clinician-agent agreement and ECI for test patients (right).
}
\label{fig7}
\end{figure}

This suggested that, while \(\pi_{\phi_A}\) and \(\pi_{\phi_B}\) generally behaved like the clinicians, their responses to certain scenarios were structurally different. Indeed, by analyzing how clinician-agent agreement varied with the Elixhauser Comorbidity Index (ECI), we confirmed that the policies responded differently to clinical complexity: \(\pi_{\phi_A}\) tended to reinforce dosing patterns typical of patients with more comorbidities, whereas \(\pi_{\phi_B}\) deviated from these patterns as the ECI increased (Fig. \ref{fig7}). 

\begin{figure}[htbp]
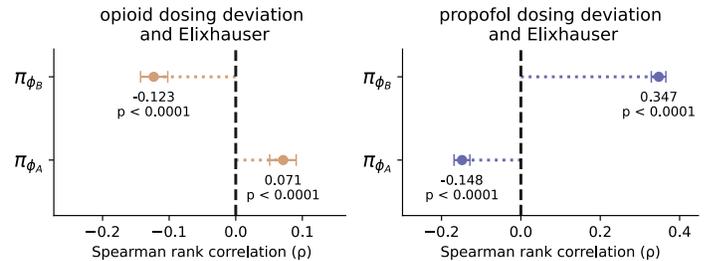

\centerline{
\includesvg[scale=0.54]{figures/opioid_deviation_and_elixhauser}
\hspace{-0.5cm}
\includesvg[scale=0.54]{figures/propofol_deviation_and_elixhauser}
}
\caption{Correlations between Elixhauser Comorbidity Index and signed deviation from the opioid and propofol doses administered by the clinicians.
}
\label{fig8}
\end{figure}

The most important difference between the two policies concerned opioid and propofol dosing. Fig.~\ref{fig8} shows that \(\pi_{\phi_A}\) responded to higher comorbidity with more aggressive, opioid-dominant strategies, while restricting propofol doses. In contrast, higher ECI led \(\pi_{\phi_B}\) to reduce opioid dosing and rely on more sustained propofol administration. Interestingly, the models never had access to diagnostic data, so both had to learn some history-based proxy for patient comorbidity.

\begin{figure}[htbp]
\vspace{-1mm}
\centerline{
\includesvg[scale=0.52]{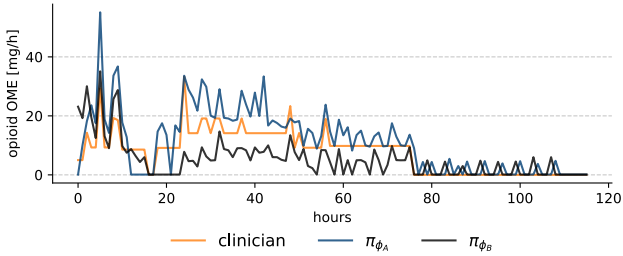}
}
\vspace{-0.5cm}
\caption{Oral morphine-equivalent (OME) dosing for a test patient with an Elixhauser Comorbidity Index of 21. When comparing the clinician, policy A (\(\pi_{\phi_A}\)) and policy B (\(\pi_{\phi_B}\)), the latter tends to prescribe lower doses.
}
\label{fig9}
\end{figure}

We found specific evidence of this behavior by studying individual high-complexity patients. Fig.~\ref{fig9} depicts the opioid doses suggested for a patient with an ECI of 21. In this scenario, the conservative dosing approach displayed by \(\pi_{\phi_B}\) aligns with clinical guidelines for fragile patients \cite{pain_guidelines}.

\begin{figure}[htbp]
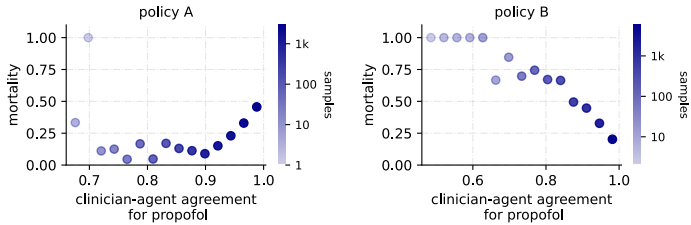

\vspace{-1mm}
\centerline{
\includesvg[scale=0.52]{figures/A_propofol_and_mortality}
\includesvg[scale=0.52]{figures/B_propofol_and_mortality}
}
\caption{Panels correspond to policy A (left) and policy B (right). Points denote propofol-dose agreement bins. Similarity to policy B is associated with lower empirical mortality, while the opposite is true for policy A. 
}
\label{fig10}
\end{figure}

Fig.~\ref{fig10} shows that, when stratifying the relationship between mortality and clinician-agent agreement by individual medications, propofol was indeed a primary contribution to \(\pi_{\phi_B}\)'s association with lower mortality. This is consistent with prior studies on opioid-propofol-based sedation and analgesia, which suggest that balanced regimens were associated with improved length of stay and mechanical ventilation outcomes when compared to opioid-dominant sedation \cite{opioid_propofol1, opioid_propofol2, opioid_propofol3}. 

%% file: sections/conclusion.tex
Data-driven RL offers a promising framework for complex clinical decision-making in the ICU. To fully realize its potential, though, it is essential to carefully consider the challenges related to goal specification and partial observability.

In line with these principles, the present work focused on investigating the risks of optimizing a single short-term goal (pain reduction) in a safety-sensitive setting. We provide evidence that pain minimization without considering long-term outcomes leads to actions associated with increased mortality in the ICU up to 30 days after discharge. This highlights a potentially critical safety limitation in previous studies, which relied on short-term clinical signals such as mean arterial pressure. In contrast, we demonstrated that explicitly valuing both pain relief and patient survival enabled the learning of data-driven RL policies associated with lower patient mortality across a large dataset with diverse comorbidity profiles. 

In obtaining these results, we extended prior research along three main dimensions. First, we modeled a richer and higher-dimensional action space, expanding from two continuous medication doses or 27 discrete options, to four continuous actions based on eight distinct drugs, including sedatives and analgesics. Second, we addressed partial observability by learning recurrent representations of the patient state while incorporating state-of-the-art offline RL methods. Third, we achieved a nearly seven-fold increase in cohort size over prior studies (which analyzed at most 6,843 admissions) by leveraging records from 47,144 ICU stays. To the best of our knowledge, this represents the most comprehensive study on data-driven RL for sedation and analgesia. 

These findings, however, should be interpreted as preliminary and observational. As a retrospective study, our results are subject to limitations like confounding by indication, selection bias, and historical clinical practice, which may itself be suboptimal. The reported effects are modest in magnitude, which is expected given the highly heterogeneous and noisy nature of real-world ICU data. In addition, they capture statistical associations, not causation. 

Confidence in these findings will be strengthened by analyzing vulnerable patient subgroups where the effects are more pronounced. Future work will also pursue off-policy evaluation with counterfactual estimators, causal modeling, a comprehensive parameter sensitivity analysis, a validation procedure using data from a multi-center cohort, and eventual prospective clinician-in-the-loop studies.

Beyond technical challenges, the use of RL in the ICU raises important ethical considerations. These systems should function strictly as decision-support tools. Thus, preserving clinician autonomy and enhancing model transparency will be essential for any future clinical translation.

%% file: sections/acknowledgment.tex
J.R. receives support from a Joan Oró research grant co-funded by Generalitat de Catalunya and the European Union (grant code: 2025 FI-1 00332). Previously, the research that led to these results was supported by a postgraduate fellowship awarded to J.R. by ”la Caixa” Foundation (ID 100010434) (fellowship code: B005785).